\documentclass[a4paper, 10pt, conference]{ieeeconf}      

\IEEEoverridecommandlockouts                              

\usepackage{graphics} 
\usepackage{epsfig} 
\usepackage{times} 
\usepackage{amsmath} 
\usepackage{amssymb}  

\newcommand\copyrighttext{%
	\scriptsize \copyright~2021 IEEE. Personal use of this material is permitted. Permission from IEEE must be obtained for all other uses, in any current or future media, including reprinting/republishing this material for advertising or promotional purposes, creating new collective works, for resale or redistribution to servers or lists, or reuse of any copyrighted component of this work in other works.}%
\newcommand\copyrightnotice{%
\begin{tikzpicture}[remember picture,overlay]
\node[anchor=south,yshift=10pt,xshift=0.25cm] at (current page.south) {{\parbox{\dimexpr\textwidth-\fboxsep-\fboxrule\relax}{\copyrighttext}}};
\end{tikzpicture}%
}

\usepackage[binary-units]{siunitx}

\usepackage{bm}
\usepackage{booktabs}
\usepackage{tikz}
\usepackage{ifthen}
\usepackage{nicefrac}
\usepackage{pgfplots}
\pgfplotsset{compat=1.7}
\usepackage{tabularx}
\usepackage{circledsteps}
\usetikzlibrary{positioning, calc, fit, shapes.geometric, decorations.pathreplacing}
\usepackage{adjustbox}
\usepackage[implicit=false, hidelinks]{hyperref}

\definecolor{plt_orange}{RGB}{255, 127, 14}
\definecolor{plt_blue}{RGB}{31, 119, 180}

\DeclareRobustCommand\sampleline[1]{%
  \tikz\draw[#1, line width=0.8pt, blue] (0,0) (0,\the\dimexpr\fontdimen22\textfont2\relax)
  -- (2em,\the\dimexpr\fontdimen22\textfont2\relax);%
}

\definecolor{landmark_red}{RGB}{230, 116, 138}
\definecolor{fill_circle}{RGB}{217, 217, 217}
\definecolor{yellow_circle}{RGB}{255, 192, 0}
\definecolor{blue_circle}{RGB}{0, 126, 240}

\DeclareRobustCommand\maptrain{\tikz[baseline=0.5ex]{\node[circle,draw=yellow_circle, line width=1pt, fill=fill_circle, inner sep=0pt, minimum size=10pt, yshift=0.2cm] {\footnotesize 1};} }
\DeclareRobustCommand\maptest{\tikz[baseline=0.5ex]{\node[circle,draw=blue_circle, line width=1pt, fill=fill_circle, inner sep=0pt, minimum size=10pt, yshift=0.2cm] {\footnotesize 2};} }
\DeclareRobustCommand\errorline{\tikz[baseline=-0.5ex]{\draw[red, line width=1pt] (0,0) -- (0.4,0);}}
\DeclareRobustCommand\errordashed{\tikz[baseline=-0.5ex]{\draw[red, dashed,line width=1pt] (0,0) -- (0.5,0);} }

\title{\LARGE \bf
Attention-based Vehicle Self-Localization with HD Feature Maps
}

\author{Nico Engel, Vasileios Belagiannis and Klaus Dietmayer
\thanks{Authors are with Institute of Measurement, Control and Microtechnology, Ulm University, Albert-Einstein-Allee 41,
89081 Ulm, Germany
        {\tt\small \{firstname.lastname\}@uni-ulm.de}.
Project Page: \url{https://github.com/engelnico/deeplocalization}. 
        }%
}

\begin{document}

\maketitle
\copyrightnotice
\thispagestyle{empty}
\pagestyle{empty}
\renewrobustcmd{\bfseries}{\fontseries{b}\selectfont}
\sisetup{detect-weight,mode=text,group-minimum-digits = 4}
\begin{abstract}
We present a vehicle self-localization method using point-based deep neural networks. Our approach processes measurements and point features, i.e. landmarks, from a high-definition digital map to infer the vehicle's pose. To learn the best association and incorporate local information between the point sets, we propose an attention mechanism that matches the measurements to the corresponding landmarks. Finally, we use this representation for the point-cloud registration and the subsequent pose regression task. Furthermore, we introduce a training simulation framework that artificially generates measurements and landmarks to facilitate the deployment process and reduce the cost of creating extensive datasets from real-world data. We evaluate our method on our dataset, as well as an adapted version of the Kitti odometry dataset, where we achieve superior performance compared to related approaches; and additionally show dominant generalization capabilities. 

\end{abstract}


\section{Introduction}
The localization of autonomous agents in an unknown environment with a high-definition (HD) digital map as prior is a key component in state-of-the-art robotic systems, including self-driving cars~\cite{thrun2002probabilistic}. It is important for other automated driving modules such as the human-vehicle interaction~\cite{wiederer2020traffic}, trajectory prediction in perception~\cite{hasan2019forecasting} and tracking~\cite{engel2018deep}. The goal is to infer the vehicle's pose, which is comprised of a position and an orientation~\cite{thrun2001robust}. Normally, the pose is estimated in the global or local coordinate system relative to the digital map, which can then be used to extract useful information from the map~\cite{gies2020extended}. 
Furthermore, the localization accuracy is expected to be around \SI{50}{\centi\meter}~\cite{levinson2007map} for \mbox{real-world applications}.

The standard vehicle self-localization approach is to rely on global navigation satellite systems (GNSS), e.g.~GPS, to obtain the pose estimate. However, they fail to meet the required localization accuracy, especially in urban environments, and suffer from multi-path effects, and blocked line-of-sight to the satellites, which further deteriorates the localization quality~\cite{wing2005consumer}. Also, the GPS signal can be combined with correction data (dGPS) and inertial measurement units (IMU) to further improve the localization accuracy, but due to the high acquisition and operation costs, it is not sustainable for mass deployment. Alternatively, one can create high-definition digital maps that contain distinct and easily recognizable high or low-level features, extracted from the on-board sensor data, such as camera, laser and radar. The sensor measurements are registered with the features from the digital map to infer the vehicle's relative pose inside the map frame. Several methods have been proposed in the field of robotics to perform the inference, ranging from simple point to point registration approaches, e.g. ICP~\cite{besl1992method}, to more sophisticated methods that utilize filtering approaches, such as the Extended Kalman-Filter (EKF) or the particle filter~\cite{montemerlo2002fastslam}. The association of the measurements to the map features, often called landmarks, is usually performed by assigning the most likely measurements to each landmark using either a heuristic or probabilistic approach. In urban environments, these algorithms suffer from erroneous associations from noise that is caused by the highly dynamic scenarios with numerous road participants~\cite{chen2020survey}.

Currently, the promising way for localization is the data-driven approach~\cite{engel2019deeplocalization, radwan2018vlocnet++}. An appropriate dataset has to be created that ideally captures most of the desired areas of operation for obtaining a model that generalizes well during deployment. However, it is not feasible to create and label datasets in hundreds of cities around the world and updating them whenever environmental changes or new scenarios emerge. We also follow the data-driven approach but propose a simulation framework to generate synthetic training data. We can train a deep neural network to perform localization without the necessity of acquiring a plethora of real-world data. Although DeepLocalization~\cite{engel2019deeplocalization} is related to our approach, it not able to learn local data relations.

We define the localization process as two tasks, namely the landmark to measurement association and the point cloud registration. For the association process, we present an attention mechanism to score each landmark assignment for subsequently learning the most probable associations. Based on the weighted representation of the measurements and landmarks, we generate local features that are used for the point cloud registration process. Then, another attention operation combines all measurements and landmarks to predict the vehicle's pose. For the inference process, we propose a GPS-based and a filter-based approach, that allows us to estimate a pose offset based on a previous pose similar to our training pipeline.
We show that employing the attention mechanism for the different tasks greatly improves the localization accuracy on different datasets compared to the related work, especially in dynamic and complex environments. Additionally, we propose a simulation framework that enables us to train the network by artificially generating landmark and measurements. We do this by designing a probabilistic model  that is inspired by the spatial occurrence of real map landmarks. The evaluation shows, that a network trained on artificial samples is able to generalize to real-world data, thus significantly reducing the resources needed to deploy our approach while still meeting the required accuracy in urban scenarios of about~\SI{50}{\centi\meter}.

\section{Related Work}
In the following, we compare and discuss related approaches for vehicle localization. We distinguish between traditional model-based approaches, such as localization methods that use filter-based algorithms, and learning-based approaches, i.e. deep neural networks. 
\subsection{Model-based approaches}
In the field of robotics, different model-based methods have been proposed to solve the task of self-localization by using a digital map as prior to estimate a pose or by simultaneously constructing a map while localizing  itself in an unknown environment, i.e. SLAM.  

Censi~\cite{censi2008icp} proposes PLICP, an improvement to the iterative closes point algorithm by introducing a point-to-line metric instead of the originally used point-to-point metric that can be used to register measurements with a feature-based map. Moreover, PLICP improves the convergence properties while also converging in a finite number of steps. Contrary to this, Fontanelli~\textit{et al.} propose a RANSAC-based localization approach using lidar measurements that is both accurate and copes with noisy measurements~\cite{fontanelli2007fast}. 
Using an Extended Kalman-Filter, Teslic \textit{et al.} introduce a localization framework  that combines wheel encoders to predict the robot's motion and laser scans to correct the robots pose by matching the measurements with a digital map~\cite{teslic2011ekf}. 

A very well-known approach by Dellaert~\textit{et al.} implements a probabilistic model by representing the robot's state space as a particle-based density~\cite{dellaert1999monte}. The Monte-Carlo Localization (MCL) is able to efficiently represent arbitrary distributions, in this case the robot's pose, while being more accurate and requiring less memory compared to related methods. Later, Thrun~\textit{et al.}~\cite{thrun2001robust} improve the algorithm by introducing two methods of generating particles in the estimation and learning a kernel density tree to enable faster sampling.
Based on the Monte-Carlo Localization, Montemerlo and Thrun develop the FastSLAM algorithm, which combines the MCL with an EKF to generate a landmark-based map while enabling a robot localization in unknown environments~\cite{montemerlo2003fastslam}. 
Additionally, Stuebler \textit{et al.} propose the RFS-MCL~\cite{stubler2015feature}, that combines the Random-Finite Set theory with a particle filter-based localization approach.
Since then, many SLAM algorithms have been proposed, e.g.~\cite{mullane2011random, deusch2015labeled}, with a heavy focus on camera-based approaches, e.g.~\cite{mur2017orb, pumarola2017pl}.
This trend of using vision-based systems can also be observed for the localization task, i.e. visual odometry (VO)~\cite{mohamed2019survey}.

\subsection{Learning-based approaches}

Besides model-based approaches, recent work focused on learning-based methods using deep neural networks. Since traditional neural networks like convolutional neural networks (CNN) or simple multi-layer perceptrons (MLP) require the input to be structured and ordered, most approaches use a vision-based system, i.e. camera images, to perform the localization task. 

Yang \textit{et al.}~\cite{yang2019sanet} propose SANet, a scene agnostic framework for camera localization, where they separate the scenes and model parameters and learn a hierarchical scene representation. Thus, SANet is independent of the scenes and can easily be deployed to online tasks, such as navigation and SLAM.
Radwan and Valada introduce a network architecture called VLocNet++~\cite{radwan2018vlocnet++}, where they combine the learning of semantics, regressing the global pose of a camera and odometry to exploit the relationships between the tasks to increase the overall performance. On the other hand, Kendall \textit{et al.} propose PoseNet~\cite{kendall2015posenet} which is a real-time localization system based on convolutional neural networks, that is designed to regress the cameras pose from a single RGB image. Similar to the localization methods using ICP variants to register two point clouds, DeepICP~\cite{lu2019deepicp} is an end-to-end trainable neural network, where a key-point detector is trained such that it focuses on stationary objects and avoids dynamic objects. This is achieved by generating corresponding points by matching possible candidates based on learned probabilities.
Wang \textit{et al.} introduce an attention-based camera relocalization system~\cite{wang2020atloc} that is robust to outliers and dynamic illumination conditions. The attention mechanism is used to learn a geometrical representation that focuses on robust features. Contrary to our approach, AtLoc is restricted to camera images only, whereas we focus on a generic feature representation that can handle multi-modal sensor measurements.
Finally, Lu \textit{et al.} propose L3-Net~\cite{lu2019l3}, a learning-based lidar localization framework that incorporates multiple network structures, such as convolutional neural networks (CNN) and recurrent neural networks (RNN), to learn local descriptors for 3D point cloud data. However, due to multiple network stages that increase the overall complexity and model size, the computational time suffers and in some cases violates our real-time requirement of about \SI{100}{\milli\second}.

A comprehensive survey on the topic of deep-learning based localization methods can be found in~\cite{chen2020survey}.

\begin{figure*}[ht]
\centering
\resizebox{\linewidth}{!} {\input{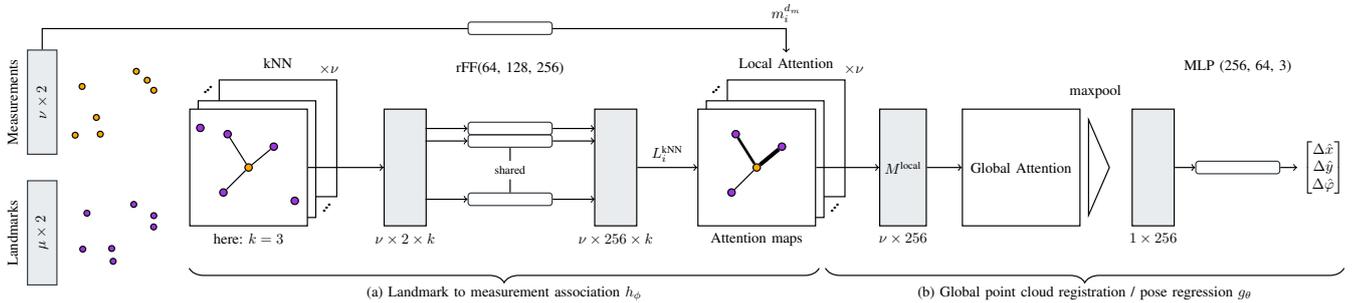}}
	\caption{Overview of our network architecture. First we group all landmarks in the vicinity of each measurement using the k-Nearest Neighbour algorithm. Then, we learn the most probable matching by employing the local attention mechanism, which weights the associations (a). Finally, we use this representation to register the measurements and the landmarks and infer a pose correction vector (b). }
	\label{fig:network_arch}
\end{figure*}

\section{Methodology}\label{sec:method}

In this section, we introduce our methodology, the attention-based model as well as our training and inference process. We assume 2D multi-modal sensor measurements as one input set, denoted by \mbox{$M = \{m_1, \dots, m_{\nu}\},~ m_{(\cdot)} \in \mathbb{R}^2$}. 
Furthermore, we consider landmarks from our digital map \mbox{$L = \{l_1, \dots, l_{\mu}\},~l_{(\cdot)} \in \mathbb{R}^2$} as the second input to our network. Landmarks are easily recognizable and static objects, e.g. trees, traffic lights, poles, that were generated from sensor measurements during the map building process, that we describe in Section \ref{seq:dataset}. Here, it is important to note that both input point sets are unordered and the cardinality of the sets, i.e. the number of measurements $\nu$ and landmarks $\mu$, is not known in advance.

\subsection{Problem Formulation}
The goal of localizing an agent is to find the relative pose $p = [x, y, \varphi]$ within a given coordinate system, e.g. the map frame. For this, we consider current multi-modal sensor measurements with the aim of matching them to the landmarks from the digital map which are in the vehicle's field of view. The matching set function is defined as following:

\begin{align}\label{eq:matching_func}
	f ( M, L ) \rightarrow [x, y, \varphi].
\end{align}
We split this set matching  problem into two subtasks: a) The landmark to measurement association, where we try to match both input point sets such that the most probable landmarks are assigned to the corresponding sensor measurements. For this, we employ the attention mechanism, which learns to score each possible landmark to measurement association. b) A global point cloud registration process in higher dimensional space using the local associations. The result of the registration process is a spatial transformation that describes the mapping of the landmarks to the measurement set, i.e. the pose prediction. In this work, we approximate the matching function \eqref{eq:matching_func} with a deep neural network, given by:
\begin{align}\label{eq:matching_func_approx}
	f (M, L) \approx g_\theta ( h_\phi (M,L) ),
\end{align}
where we denote the set of learnable parameters by $\theta, \phi$. Furthermore, the landmark to measurement association process is defined as $h_\phi (\cdot)$ and the point cloud registration together with the subsequent pose regression is combined as $g_\theta (\cdot)$.

\subsection{Attention-based model}
The measurements $M$ and the landmarks (features) $L$ serve as input to our network, which is visualized in Fig.~\ref{fig:network_arch}. In the first step of the landmark to measurement association process, see Fig.~\ref{fig:network_arch}a), we employ a k-Nearest Neighbour (kNN) search to find the $k$ closest landmarks to each measurement. We perform the kNN to take into account the spatial relationship between measurements and landmarks: Measurements ideally originate from landmarks, therefore we only consider landmarks in the vicinity of the measurements for the association process.
Then, for every measurement we calculate the Euclidean distance of the $k$ associated landmarks and transform the result into higher dimensional space $d_m = 256$ using a row-wise feed forward (rFF) network. A rFF is a feed-forward network that receives only one point as input but shares its weight with all subsequent points~\cite{qi2017pointnet++}. Then we employ a multi-head attention module~\cite{vaswani2017attention}, as it can capture context and higher order dependencies of point sets~\cite{engel2020point}. 
Furthermore, we leverage the fact that the attention mechanism scores the input sets, thus we learn a weighting of the most probable association.
For that reason, we define the attention function $\mathcal{A}$ that describes a mapping of $N$ queries $Q \in \mathbb{R}^{N\times d}$ and $N_k$ key-value pairs $K \in \mathbb{R}^{N_k \times d}$, $V \in \mathbb{R}^{N_k \times d}$ to the output space $\mathbb{R}^{N \times d}$ as follows

\begin{align}\label{eq:attention}
	\mathcal{A}(Q,K,V) = \sigma \left ( \frac{QK^T}{\nicefrac{1}{\sqrt{d}}} \right )V,
\end{align}
which is also known as scaled dot product attention $\mathcal{A}(Q,K,V) :  \mathbb{R}^{N\times d_k},  \mathbb{R}^{N_k\times d},  \mathbb{R}^{N_k\times d} \rightarrow \mathbb{R}^{N\times d}$~\cite{vaswani2017attention}. The activation function $\sigma(\cdot)$ that also denotes the aforementioned learned score is usually given by the softmax function 

\begin{align}
	\sigma_\text{softmax}(x_i) = \frac{\exp(x_i)}{\sum_j \exp(x_j)},
\end{align}
with $\sigma(\cdot): \mathbb{R}^{N\times d},\mathbb{R}^{N_k\times d} \rightarrow \mathbb{R}^{N\times N_k}$. Instead of performing a single attention operation, we follow the ideas of~\cite{vaswani2017attention} and employ multi-head attention, where the queries, keys and values are first linearly projected $h$ times using independent feed-forward networks to incorporate spatial relations in different subspaces. Then, the attention function \eqref{eq:attention} is applied in parallel to each of the $h$ projections and the result is concatenated and linearly projected again. The operation is described by:

\begin{align}
    \text{Multihead}(Q,K,V) = (\text{head}_1 \oplus ... \oplus \text{head}_h) W^O,
\end{align}
where $\text{head}_i = \mathcal{A}(QW^Q_i, KW^ K_i, VW^V_i)$ with learnable parameter matrices $W^Q_i \in \mathbb{R}^{d \times d}$, $W^K_i \in \mathbb{R}^{d \times d}$,  $W^V_i \in \mathbb{R}^{d \times d}$ and $W^O \in \mathbb{R}^{d \times d}$.
The $\oplus$ operation denotes matrix concatenation. Finally, Vaswani \textit{et al.} define the multi-head attention block that consists of the attention operation \eqref{eq:attention} and residual connections followed by layer normalization~\cite{ba2016layer} as follows:

\begin{align}\label{eq:mha}
    \mathcal{A}^{\text{MH}} (X,Y) = \text{LayerNorm}(S + \text{rFF}(S)),
\end{align}
where $\mathcal{A}^{\text{MH}} : \mathbb{R}^{N\times d},\mathbb{R}^{N_k \times d} \rightarrow  \mathbb{R}^{N\times d}$. and the sublayer $S$ is defined as $S = \text{LayerNorm}(X + \text{Multihead}(X,Y,Y))$. Furthermore, $X$ and $Y$ denote arbitrary input sets. In the following we set $d = d_m = 256$.

As visualized in Fig.~\ref{fig:network_arch}, we employ local attention for subtask (a) and define it as follows:
\begin{align}\label{eq:local_attn}
	\mathcal{A}^\text{local}_i := \mathcal{A}^{\text{MH}}(m_i^{d_m}, L_i^\text{kNN}),\, i = 1, \dots, \nu,
\end{align}
where we take each measurement $m_i$, project it to model dimension $d_m$ using the rFF depicted in Fig.~\ref{fig:network_arch} and apply the multi-head attention mechanism \eqref{eq:mha} with the $k$ associated landmarks from the kNN algorithm. Thus, we have \mbox{$m_i^{d_m} \in \mathbb{R}^{1\times d_m}$}, \mbox{$L_i^\text{kNN} \in \mathbb{R}^{k \times d_m}$} and $\mathcal{A}^\text{local}_i \rightarrow \mathbb{R}^{1\times d_m}$. By employing the local attention operation \eqref{eq:local_attn} we generate latent features for every measurement that contain a weighted representation of all nearby landmarks, see Equation \eqref{eq:attention}. Thus, our network is able to learn the most probable landmark to measurement association which is visualized by the weight of the connection in the attention maps in Fig.~\ref{fig:network_arch}. After applying Equation~\eqref{eq:local_attn} to each measurement and concatenating the result, we obtain the local feature matrix $M^\text{local} \in \mathbb{R}^{\nu \times d_m}$. Then, we employ another self-attention operation for subtask (b) to aggregate global information

\begin{align}\label{eq:global_attn}
    \mathcal{A}^\text{global} := \mathcal{A}^{\text{MH}}(M^\text{local}, M^\text{local}),
\end{align}
with $\mathcal{A}^\text{global}: \mathbb{R}^{\nu \times d_m},\mathbb{R}^{\nu \times d_m} \rightarrow \mathbb{R}^{\nu \times d_m}$. 
The global attention mechanism~\eqref{eq:global_attn} relates the local features of the matched landmarks and measurements against each other, allowing the network to learn the point set registration process and capture higher order dependencies.
Afterwards, a maxpooling operation~\cite{qi2017pointnet} follows to produce global features of fixed length that are invariant to input point permutations and arbitrary input set cardinality. Here, it is important to note that the attention operation itself is invariant to input point permutations as well~\cite{vaswani2017attention}.
Finally, we infer the vehicle's pose using the learned global features with a simple feed forward output head as shown in Fig.~\ref{fig:network_arch}.

\subsection{Model training}\label{sec:training}

For training, we follow the same protocol as in~\cite{engel2019deeplocalization} to treat the problem as a regression task~\cite{belagiannis2015robust} and rely on the ground-truth pose from our dGPS system as a starting point. Then, all landmarks in the vehicle's field of view (FoV) are loaded from the digital map and are  transformed from UTM coordinates into the vehicle's frame using the dGPS pose. At that point, both inputs, namely the landmarks and measurements, are available in the same coordinate system. To imitate real-world conditions during training, we additionally add a small translation and rotation to all landmarks by sampling both a position and a rotation offset from a uniform distribution $\mathcal{U}$ on the interval $[-\sigma, \sigma]$. Thus, we systematically simulate the inaccurate GPS measurement $p_\text{GPS}$. The advantage of this method is that it is no longer necessary to determine the global pose, but instead we infer a synthetically the generated pose offset. Therefore, Equation~\eqref{eq:matching_func} becomes
\begin{align}
    f(M,L) \rightarrow [\Delta x, \Delta y, \Delta \varphi],
\end{align}
with the pose offset $\Delta p = [\Delta x, \Delta y, \Delta \varphi]$.  The predicted global pose $\hat{p}$ can then be easily obtained by 
\begin{align}
    \hat{p} = p_\text{GPS} - \Delta \hat{p},
\end{align}
where the prediction of the network is denoted by $\Delta \hat{p}$.
This method allows us to generate new training samples every epoch, as the randomly sampled pose offset directly serves as the training label. Furthermore, it simplifies the training of the network because we found that inferring a small offset to be more numerically stable in the optimization process compared to using the global UTM coordinate system.
Since the rotation and translation offset are given in different units, i.e. $\si{\radian}$ and $\si{\meter}$, we train our network with the same loss function that learns a weighting factor for each of the loss parts, as proposed in~\cite{kendall2018multi, engel2019deeplocalization}. In particular, we employ the L2-Loss for each of the pose component predictions
\begin{subequations}
\begin{align}
		L_\text{tran} &= \mathbb{E}[(\Delta \hat{x} - \Delta x)^2] + \mathbb{E}[(\Delta \hat{y} - \Delta y)^2], \\ 
		L_\text{rot} &= \mathbb{E}[(\Delta \hat{\varphi} - \Delta \varphi)^2],
	\end{align}
\end{subequations}
where $L_\text{tran}$ is the translation loss and $L_\text{rot}$ the rotation loss. For the total multi-task loss, we combine the loss terms 
\begin{align}\label{eq:geom_loss}
	L_\text{multi} = L_\text{tran} e^{-s_\text{tran}} + s_\text{tran} + L_\text{rot} e^{-s_\text{rot}} + s_\text{rot},
\end{align}
where $s_\text{tran} = \log \sigma_\text{tran}^2$, $s_\text{rot} = \log \sigma_\text{rot}^2$. 
Following the ideas from Kendall \textit{et al.}, $\sigma_\text{tran}$,$\sigma_\text{rot}$ are the homoscedastic uncertainties, i.e.~learnable parameters for weighting each loss function~\cite{kendall2018multi}. 

\begin{figure}
	\centering
	\resizebox{\linewidth}{!}{
	\begin{tikzpicture}

	\node (minput) at (0,0) {$M^\text{veh}$};
	\node[] (linput) at (0, -1) {$L^\text{UTM}$};
	
	\node at ($(minput) + (2, 1.0)$) {\textbf{a) GPS-based Inference:}};
	
	\node[draw, rectangle, align=center, line width=0.8pt, minimum width=2.5cm, minimum height=1.5cm, rounded corners] (network) at ($(minput)!0.5!(linput) + (5, 0)$) {Attn-based \\ Localization}; 
	
	\node[inner sep=0pt] (vehtrans) at ($(linput.east)!0.5!(network.west) - (0, 0.24)$) {$\bigotimes$}; 
	
	\draw [->, line width = 0.8pt] (minput) -- (minput -| network.west);
	\draw [->, line width = 0.8pt] (linput) -- (linput -| vehtrans.west);
	\draw [->, line width = 0.8pt] (vehtrans.east) -- node [above] {$L^\text{veh}_\text{GPS}$} (vehtrans.east -| network.west);

	\node (phat) at ($(network) + (2.5,0)$) {$\Delta \hat{p}$};
	\node [draw, line width= 0.8pt, inner sep=8pt] (pgps) at ($(network.south) + (0, -0.8)$) {$p_\text{GPS}$};
	\node (p) at ($(phat) + (2.5,0)$) {$\hat{p}$};

	\node[inner sep=0pt] (globalp) at ($(phat)!0.5!(p)$) {$\bigoplus$};

	\draw[->, line width = 0.8pt] (pgps) -| (vehtrans);
	\draw[->, line width = 0.8pt] (pgps) -| (globalp);
	
	\draw[->, line width= 0.8pt] (network) -- (phat);
	\draw[->, line width= 0.8pt] (phat) -- node[above] {\footnotesize{$-$}} (globalp);
	\draw[->, line width= 0.8pt] (globalp) -- (p);

	\node (minput_filter) at ($(minput) - (0, 4.5)$) {$M^\text{veh}_t$};
	\node[] (linput_filter) at ($(minput_filter) - (0, 1)$) {$L^\text{UTM}$};
	
	\node at ($(minput_filter) + (2, 1.0)$) {\textbf{b) Filter-based Inference:}};
	
	\node[draw, rectangle, align=center, line width=0.8pt, minimum width=2.5cm, minimum height=1.5cm, rounded corners] (network_filter) at ($(minput_filter)!0.5!(linput_filter) + (5, 0)$) {Attn-based \\ Localization}; 
	
	\node[inner sep=0pt] (vehtrans_filter) at ($(linput_filter.east)!0.5!(network_filter.west) - (0, 0.24)$) {$\bigotimes$}; 
	
	\draw [->, line width = 0.8pt] (minput_filter) -- (minput_filter -| network_filter.west);
	\draw [->, line width = 0.8pt] (linput_filter) -- (linput_filter -| vehtrans_filter.west);
	\draw [->, line width = 0.8pt] (vehtrans_filter.east) -- node [above] {$L^\text{veh}_{t-1}$} (vehtrans_filter.east -| network_filter.west);

	\node (phat_filter) at ($(network_filter) + (2.5,0)$) {$\Delta \hat{p}_t$};
	\node [draw, line width= 0.8pt, inner sep=8pt] (pgps_filter) at ($(network_filter.south) + (0, -0.8)$) {$p_{t-1}$};
	\node (p_filter) at ($(phat_filter) + (2.5,0)$) {$\hat{p}_t$};

	\node[inner sep=0pt] (globalp_filter) at ($(phat_filter)!0.5!(p_filter)$) {$\bigoplus$};

	\draw[->, line width = 0.8pt] (pgps_filter) -| (vehtrans_filter);
	\draw[->, line width = 0.8pt] (pgps_filter) -| (globalp_filter);
	
	\draw[->, line width= 0.8pt] (network_filter) -- (phat_filter);
	\draw[->, line width= 0.8pt] (phat_filter) -- node[above] {\footnotesize{$-$}} (globalp_filter);
	\draw[->, line width= 0.8pt] (globalp_filter) -- (p_filter);

	\node[draw, rectangle, align=center, line width=0.8pt, minimum width=1.5cm, minimum height=1.5cm, rounded corners] (ekf) at ($(p_filter) + (1.5,0)$) {EKF \\ (CTRV)};

	\node (pekf_filter) at ($(ekf) + (2,0)$) {$\hat{p}_t^\text{EKF}$};
	\draw[->, line width= 0.8pt] (p_filter) -- (ekf);
	\draw[->, line width= 0.8pt] (ekf) -- (pekf_filter);
	
	\draw[->, line width= 0.8pt, dotted] (pekf_filter.south) --  ($(pekf_filter.south) + (0, -2.0)$)  -| node[above, midway, xshift=5cm] {update}  (pgps_filter.south);
	
\end{tikzpicture}
	}
	\caption{Overview of the proposed inference methods.}
	\label{fig:inference}
\end{figure}
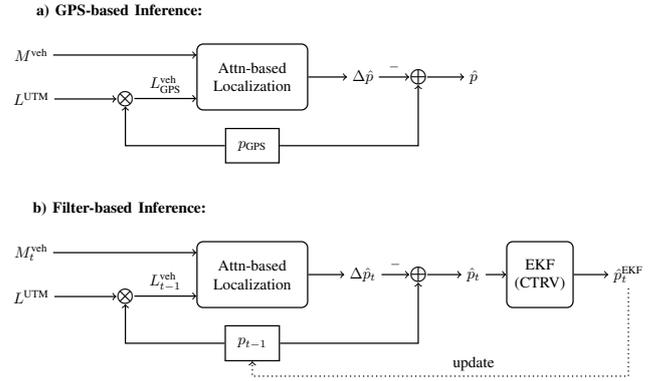

\subsection{Inference}

Similar to~\cite{engel2019deeplocalization}, we present two different inference approaches, which are also visualized in Fig.~\ref{fig:inference}.

\textbf{GPS-based inference} is the default inference configuration that resembles the training process from Section~\ref{sec:training}. Again, we load all landmarks from the digital map that are in the vehicle's field of view $L^\text{UTM}$ and transform them into the vehicle's coordinate system $L^\text{veh}_\text{GPS}$ using the noisy GPS measurement $p_\text{GPS}$, which is usually supplied in global coordinates, e.g. UTM. Since the pose that is used for transforming the landmarks to the vehicle coordinate system is noisy and inaccurate, it induces a small shift and rotation to the landmarks, which resembles the synthetic and randomly sampled pose offset that is applied to the ground-truth pose in the training process.
As mentioned above, the goal is to infer a pose correction vector $\Delta \hat{p}$ that is applied to the initial pose estimate $p_\text{GPS}$ in order to obtain the global pose $\hat{p}$, see Fig.~\ref{fig:inference}~a). This inference approach can directly be applied without further adjustments and is our recommended configuration when a commercially available GPS sensor with noisy measurements is installed.

\textbf{Filter-based inference} extends our system architecture with an Extended Kalman-Filter in order to obtain a temporal filtered and smooth pose estimate. To highlight the incorporation of the time domain, we slightly change our notation as shown in Fig.~\ref{fig:inference}~b). The algorithm requires an previous pose estimate $p_{t-1}$ which can either be supplied by a commercially available and noisy GPS system for initialization or from the previous time step $t-1$. Similar to the GPS-based inference, we use $p_{t-1}$ to transform the landmarks from the digital map $L^\text{UTM}$ to the vehicle coordinate system $L^\text{veh}$ and use it as input to our network together with the current measurements. Since landmarks that are transformed with the previous pose estimate together with current measurements are used as input to our network, we again obtain a small shift and rotation due to the vehicle's motion. 
The correction output $\Delta \hat{p}_t$ is applied to the previous pose estimate $p_{t-1}$, which is then used as measurement input to the Extended Kalman-Filter with a constant turn rate and velocity motion model (CTRV). 
The output $\hat{p}_t^\text{EKF}$ is a smoothed estimate of the global pose. In the next time step, we use this estimate as the initial pose $p_{t-1}$. The advantage of the filter-based inference approach is that it requires only one GPS measurement for initialization and can then be used as a stand-alone localization method. Furthermore, in~\cite{engel2019deeplocalization} we show that the computational overhead of the EKF is negligible.

\begin{figure}[t!]
	\centering
	\includegraphics[width=0.8\linewidth]{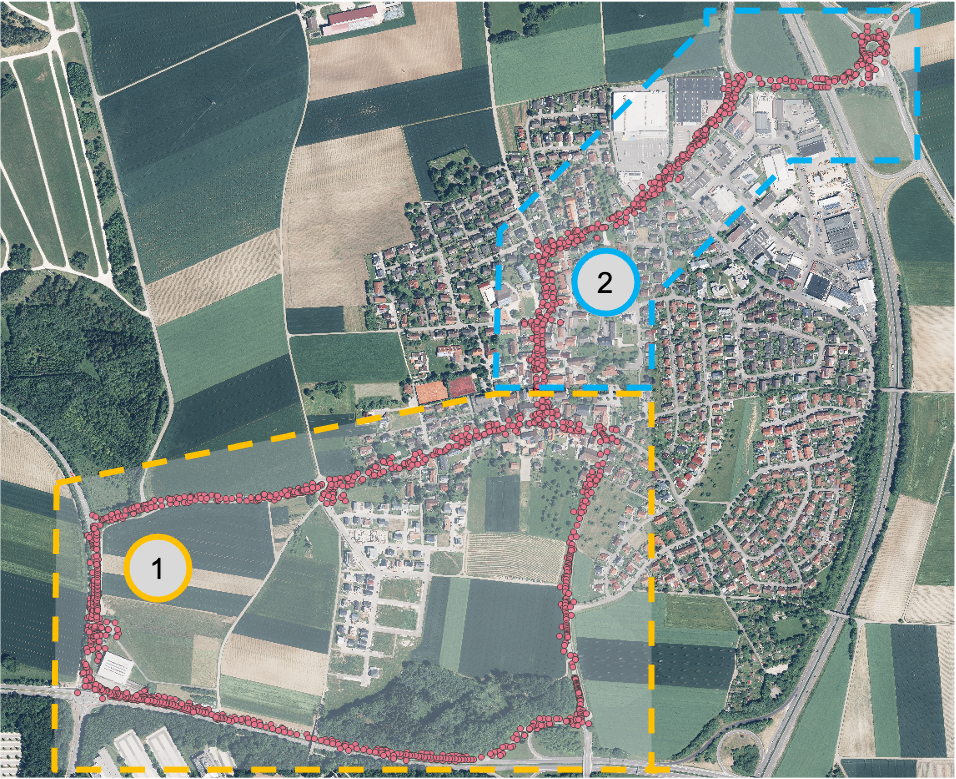}
	\caption{Our test track in Ulm-Lehr. Landmarks are shown as red dots (\textcolor{landmark_red}{$\bullet$}) and we visualize the alternative \mbox{Train / Test} split as \maptrain / \maptest , respectively. }
	\label{fig:geo_map}
\end{figure}

\section{Dataset}\label{seq:dataset}

In this section, we introduce our dataset which consists of a high-definition (HD) digital map with point features, i.e. landmarks, and multiple recordings with sensor measurements on our test track in Ulm, Lehr, that is shown in Fig.~\ref{fig:geo_map}. The test track is about \SI{6}{\kilo\meter} long with urban and rural roads, intersections, roundabouts and merging lanes. For the measurement dataset, we recorded multiple runs in November 2018 using our autonomous vehicle which is equipped with a camera, radar and laser sensors, as well as a dGPS system that provides high precision pose information that we use for the training process and for evaluation purposes. Additionally, we employ a simple pre-processing pipeline where we cluster the raw measurements to obtain more stable and reliable point features. For this, we cluster the laser and radar measurements using the density-based DBSCAN~\cite{ester1996density} and for camera images, we extract features using the maximally stable extremal region algorithm (MSER)~\cite{matas2004robust}. For every measurement time step, we additionally record the vehicle's high-precision pose using the dGPS system. Finally, for training our network we introduce two train / test splits: 1) a uniform distribution, in which we use six complete runs for the training split and two additional runs for the test split, and 2) a spatial split visualized as \maptrain / \maptest in Fig.~\ref{fig:geo_map}, to demonstrate the generalization capabilities of our approach in unseen environments. Our high-definition (HD) map was generated one year before the measurement dataset to incorporate a diverse data distribution that closely resembles the dynamic environment prevalent in most urban scenarios. As mentioned above, our map contains generic landmarks (features) that are created from our pre-processed sensor measurements. During our map building process, we classify a single measurement as static and easily recognizable when it is seen multiple times on different runs. For that, a Bernoulli-filtering approach is employed as proposed by Stuebler~\textit{et al.}~\cite{stubler2017continuously}, that assigns an existence probability to every landmark candidate. Finally, we only select landmarks that have a high existence probability to build our map. In total, the digital map consists of $3860$ landmarks and has a size of only \SI{600}{\kilo\byte}. Furthermore, the landmarks are stored in the Universal Transverse Mercator (UTM) coordinate system, while the measurements are stored in the vehicle frame that has its origin at the center of the rear axle.
An in-depth explanation of the map creation process and the pre-processing pipeline can be found in~\cite{stubler2017continuously, engel2019deeplocalization}.
Additionally, we use the Kitti odometry dataset~\cite{Geiger2012CVPR} to evaluate our approach on a public dataset. Since our approach relies on a single feature for each object, we only use the provided camera images and extract landmarks for the digital map and measurements using the MSER algorithm as described above.

\begin{figure*}[t]
	\centering
	\includegraphics[width=0.9\textwidth]{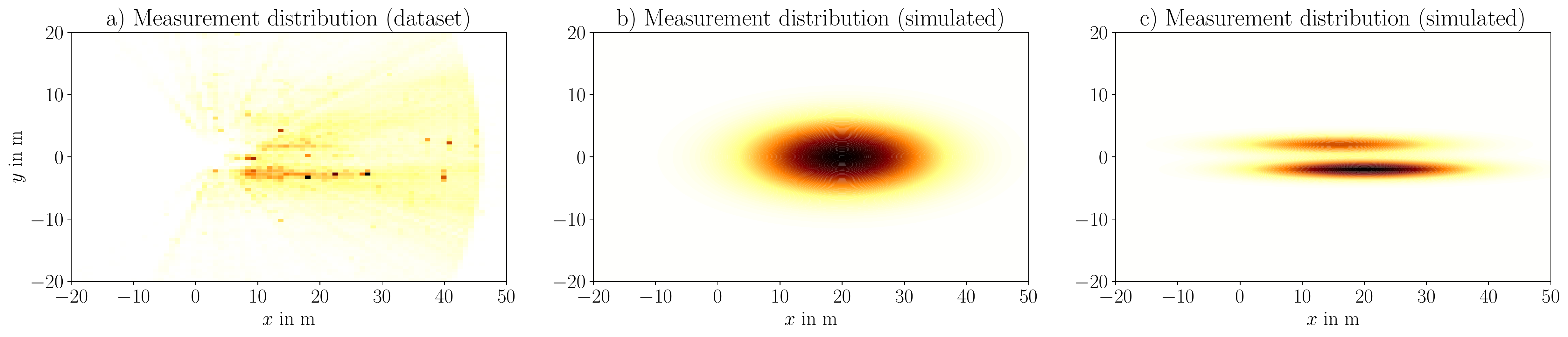}
	\caption{Measurement distribution from our dataset (left) and two possible approaches for the simulated measurement model using 2D Gaussian distributions (center and right). Darker patches indicate areas where measurements appear more frequently. }
	\label{fig:sim_meas_dist}
\end{figure*}

\section{Training Simulation Framework}\label{sec:sim_framework}

Besides our approach, we present a training simulation framework that allows us to synthetically generate training data without the need of recording real world measurements. Ideally, our landmark-based localization method can be deployed in a variety of environments and scenarios. This requires a diverse and extensive training dataset that enables the network to generalize to unseen data points. However, creating and maintaining a large-scale dataset in multiple urban and rural areas is both expensive and labor intensive. Instead, we propose a training simulation framework, which is based on real-world data. Since static objects are usually located at the side of the road, e.g. traffic signs, poles and trees, the landmarks are distributed mostly along the longitudinal axis, as we show in Fig.~\ref{fig:sim_meas_dist}~a), where darker areas indicate a more frequent occurrence of landmarks in the vehicle's field of view. Based on the measurement distribution of our dataset, we model the location using a multivariate normal distribution with mean $\mu \in \mathbb{R}^2$ and covariance matrix $\Sigma \in \mathbb{R}^{2 \times 2}$ given by:
\begin{align}
    \mathcal{N}(\mu, \Sigma) = \frac{1}{\sqrt{(2\pi)^2 |\Sigma|}} \exp \left ( - \frac{1}{2} (x-\mu)^T \Sigma^{-1} (x-\mu)  \right ).
\end{align}
 In Fig.~\ref{fig:sim_meas_dist}~b) and c), we demonstrate two possibilities of modelling the measurement distribution. In b) we set \mbox{$\mu = \begin{bmatrix} 
    20 & 0
 \end{bmatrix}$} and $\Sigma = \text{diag}(100, 15)$. To better incorporate the fact that the measurements occur on the side of the road, we use a Gaussian mixture $\mathcal{N}(\mu, \Sigma) = \mathcal{N}_1 (\mu_1, \Sigma_1) + \lambda_2  \mathcal{N}_2 (\mu_2, \Sigma_2)$, with $\mu_1 = \begin{bmatrix}
 20 & -2
 \end{bmatrix}$, $\mu_2 = \begin{bmatrix}
 20 & 2
 \end{bmatrix}$, $\Sigma_1 = \Sigma_2 = \text{diag}(120, 1)$ and $\lambda_2 = 0.6$.
The resulting Gaussian is visualized in Fig.~\ref{fig:sim_meas_dist}~c). Each training step, we randomly draw $\nu$ measurements from the Gaussian $M \sim N(\mu, \Sigma)$, where again $\nu$ is a random variable $\nu \sim \mathcal{U}(\nu_{\min}, \nu_{\max} )$. Considering ideal sensors that detect each landmark in the vehicle's field of view, the set of measurements and landmarks are equal $M = L$, thus we duplicate all our synthetically sampled measurements. However, we identified three effects that deteriorate the quality of the input data: 1) \textit{Clutter measurements} did not originate from a landmark either due to dynamic objects or a faulty sensor. 2) \textit{Missed detections} are landmarks that were not seen by any sensor because the landmark was obscured or, again, caused by a faulty sensor. And finally, 3) \textit{measurement noise} that causes an inaccurate landmark localization. We model these effects with a Poisson distribution $P(k;\lambda) = (\lambda^k \exp(-\lambda))/(k!)$, which simulates the number of events with the mean occurrence rate $\lambda$. In our case, we add clutter measurements with a mean clutter rate of $\lambda_\text{clutter}$ and we delete measurements with mean miss rate $\lambda_\text{miss}$.
Finally, we add noise to the remaining measurements by sampling from a uniform distribution $\mathcal{U}(-\sigma_\text{noise},\sigma_\text{noise})$. 

To summarize, we first generate $\nu$ measurements from the spatial distribution, visualized in Fig.~\ref{fig:sim_meas_dist}. These measurements are duplicated and then used as landmarks. To simulate real-word scenarios and environments, we deteriorate the measurements by applying multiple effects, such as clutter, missed detections and measurement noise. The modified measurements and landmarks form the input to our network, following the training process from Section~\ref{sec:training}. 

\section{Experiments and Evaluation}
Here, we present the experiments and the evaluation results. In particular, we perform real-world experiments on our Ulm-Lehr datasets as well as the adapted version of the Kitti odometry dataset~\cite{Geiger2012CVPR}. Then, we evaluate our network architecture using the simulation framework as introduced in Section~\ref{sec:sim_framework} and compare the localization accuracy by changing the percentage of simulated and real-world training data. Our network is implemented using \textit{PyTorch}~\cite{paszke2019pytorch} and we perform all experiments on a \textit{Nvidia Geforce 2080Ti}. Furthermore, we set $k=8$ as we found it to capture all landmarks that are in the vicinity of each measurement.

\subsection{Real-World Experiments}
The results of our real-world experiments are shown in Table~\ref{tab:real_experiments}, where we
report the root mean square error (RMSE).
First, we compare the localization accuracy of the GPS-based inference with our prior work~\cite{engel2019deeplocalization} for different GPS noise parameters, which are denoted by $\sigma_x, \sigma_y$ and $\sigma_\varphi$ in \mbox{Table~\ref{tab:real_experiments}~a) - c)}. 
Our approach shows impressive improvements in every experiment we conducted both on our own and the Kitti odometry dataset. In some cases, we improve the localization accuracy up to $54~\%$. Furthermore, DeepLocalization~\cite{engel2019deeplocalization} has about $1.8$~M learnable parameters with an inference time of about \SI{2}{\milli\second}. Due to the attention mechanism, the complexity of our network increases to $8.4$~M learnable parameters with an inference time of \SI{26}{\milli\second}, which still meets our real-time requirements.

Next, we compare the filter-based inference with two baseline approaches (ICP~\cite{besl1992method} and an EKF), as well as related state-of-the-art model-based localization methods in \mbox{Table~\ref{tab:real_experiments}~d) - e)}. Here, it is important to note that we modified the FastSLAM~\cite{montemerlo2002fastslam} and the PHDSlam~\cite{mullane2010rao} to only include the localization algorithm, as the digital map is already provided by our dataset. Similar to the GPS-based inference, we achieve the best localization accuracy by a wide margin on the Ulm-Lehr dataset as well as the Kitti odometry dataset compared to the related work with a mean accuracy of about \SI{0.16}{\meter} - \SI{0.18}{\meter} for the position and $1.4^\circ$ for the orientation.

\begin{figure*}[t]
	\centering
	\includegraphics[width=0.91\textwidth]{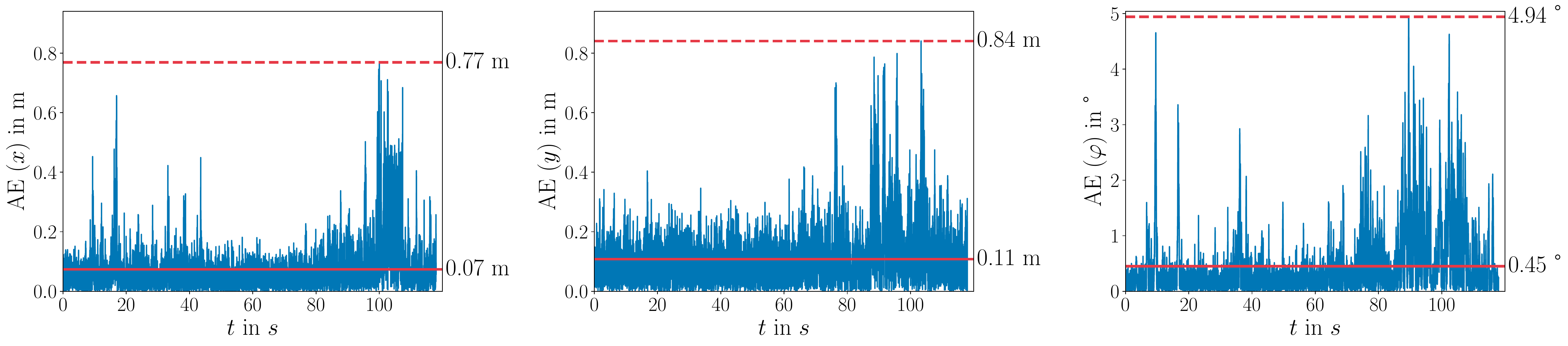}
	\caption{Here, we report the absolute localization errors (AE) on our test track in Ulm-Lehr, the RMSE (\errorline) and the maximum error (\errordashed) for the filter-based inference. }
	\label{fig:rmse_plot}
\end{figure*}

Additionally, we evaluate the GPS-based inference with noise parameters $\sigma_x, \sigma_y = 1~\si{\meter}, \sigma_\varphi = 4^\circ$ on our alternative train/test split, visualized as  \maptrain / \maptest in Fig.~\ref{fig:geo_map}, to show the generalization capabilities of our approach. With the uniform split our network achieves a localization accuracy of \SI{0.20}{\meter} to \SI{0.23}{\meter} and an orientation accuracy of $1.7^\circ$. When we train the network with the spatially divided dataset \mbox{\maptrain / \maptest}, the localization accuracy only drops to \SI{0.29}{\meter} and \SI{0.33}{\meter} for the $x$ and $y$ component, respectively. We achieve an orientation accuracy of about $1.9^\circ$. These results show, that even though the network has never been trained on the area marked as \mbox{\maptest}, our approach is able to localize in an unknown environment and achieve an accuracy that meets the requirement for urban scenarios. 
Finally, the localization accuracy of an exemplary \SI{2}{\minute} drive on our test track is shown in Fig.~\ref{fig:rmse_plot}. We additionally highlight the RMSE as well as the maximum error for each pose component.

\begin{table}[t!]
\caption{Results of Real-World Experiments.}
\label{tab:real_experiments}
\renewcommand{\arraystretch}{1.4}
\begin{tabularx}{\linewidth}{@{}lXXXXXX@{}}
\toprule
        & \multicolumn{3}{c}{Ulm-Lehr} & \multicolumn{3}{c}{Kitti Odometry~\cite{Geiger2012CVPR}}              \\
Method & $x$       & $y$       & $\varphi$      & $x$         & $y$         & $\varphi$   \\ \midrule

\multicolumn{7}{l}{\textbf{a) GPS-based Inference:}  $\sigma_x, \sigma_y = 2~\si{\meter}, \sigma_\varphi = 10^\circ$} \\
  DeepLoc~\cite{engel2019deeplocalization}     &   $0.77~\si{\meter}$      &   $0.70~\si{\meter}$      &   $2.5^{\circ}$       &  $0.84~\si{\meter}$         &    $0.82~\si{\meter}$             &     $3.1^{\circ}$          \\
  
  Attn-based (ours) & \bfseries \SI[detect-weight = true]{0.41}{\metre} & \bfseries \SI[detect-weight = true]{0.51}{\metre} & \bfseries \SI[detect-weight = true]{1.7}{\degree} &  \bfseries \SI[detect-weight = true]{0.49}{\metre} & \bfseries \SI[detect-weight = true]{0.55}{\metre} & \bfseries \SI[detect-weight = true]{2.1}{\degree}\\

  \multicolumn{7}{l}{\textbf{b) GPS-based Inference:} $\sigma_x, \sigma_y = 1~\si{\meter}, \sigma_\varphi = 4^\circ$} \\
  DeepLoc    &    $0.44~\si{\meter}$     &    $0.37~\si{\meter}$     &   $1.3^{\circ}$       &           $0.48~\si{\meter}$ &    $0.44~\si{\meter}$             &   $2.3^{\circ}$            \\
   
  Attn-based (ours) & \bfseries \SI[detect-weight = true]{0.20}{\metre} & \bfseries \SI[detect-weight = true]{0.23}{\metre} & \bfseries \SI[detect-weight = true]{0.9}{\degree} &  \bfseries \SI[detect-weight = true]{0.29}{\metre} & \bfseries \SI[detect-weight = true]{0.32}{\metre} & \bfseries \SI[detect-weight = true]{1.7}{\degree}\\

  \multicolumn{7}{l}{\textbf{c) GPS-based Inference:} $\sigma_x, \sigma_y = 0.5~\si{\meter}, \sigma_\varphi = 2^\circ$} \\
  DeepLoc     &    $0.27~\si{\meter}$     &    $0.23~\si{\meter}$     &   $0.8^{\circ}$       &           $0.35~\si{\meter}$ &    $0.33~\si{\meter}$             &    $1.4^{\circ}$           \\

  Attn-based (ours) & \bfseries \SI[detect-weight = true]{0.17}{\metre} & \bfseries \SI[detect-weight = true]{0.18}{\metre} & \bfseries \SI[detect-weight = true]{0.6}{\degree} &  \bfseries \SI[detect-weight = true]{0.21}{\metre} & \bfseries \SI[detect-weight = true]{0.24}{\metre} & \bfseries \SI[detect-weight = true]{1.1}{\degree}\\

   \midrule
    \multicolumn{7}{l}{\textbf{d) Related approaches}} \\
 ICP~\cite{besl1992method} & $1.17~\si{\meter}$ & $1.46~\si{\meter}$ & $4.9^{\circ}$ & $1.91~\si{\meter}$ & $1.84~\si{\meter}$ & $6.1^{\circ}$ \\
 EKF + GPS & $0.59~\si{\meter}$ & $0.54~\si{\meter}$ & $6.5^{\circ}$  & $1.07~\si{\meter}$ & $1.12~\si{\meter}$ & $6.5^{\circ}$  \\
 FastSlam~\cite{montemerlo2002fastslam} & $0.34~\si{\meter}$ & $0.32~\si{\meter}$ & $2.1^{\circ}$ & $0.73~\si{\meter}$ & $0.77~\si{\meter}$ & $2.9^{\circ}$  \\
 PHDSlam~\cite{mullane2010rao} & $0.30~\si{\meter}$ & $0.32~\si{\meter}$ & $1.7^{\circ}$ & \multicolumn{3}{c}{---} \\
 RFS-MCL~\cite{stubler2015feature} & $0.28~\si{\meter}$ & $0.26~\si{\meter}$ & $1.9^{\circ}$ & \multicolumn{3}{c}{---}  \\
  \multicolumn{7}{l}{\textbf{e) Filter-based Inference}} \\
 DeepLoc + EKF & $0.27~\si{\meter}$ & $0.24~\si{\meter}$ & $0.8^{\circ}$ & $0.45~\si{\meter}$ & $0.44~\si{\meter}$ & $2.1^{\circ}$  \\
Attn-based + EKF (ours) &     \bfseries \SI[detect-weight = true]{0.18}{\metre} & \bfseries \SI[detect-weight = true]{0.16}{\metre} &  \bfseries \SI[detect-weight = true]{0.6}{\degree} & \bfseries \SI[detect-weight = true]{0.31}{\metre} & \bfseries \SI[detect-weight = true]{0.25}{\metre} & \bfseries \SI[detect-weight = true]{1.4}{\degree}     \\ 
 \bottomrule
\end{tabularx}

\end{table}

\begin{table}[t!]
\caption{Results of Simulated Experiments.}
\label{tab:sim_experiments}
\renewcommand{\arraystretch}{1.4}
\begin{tabularx}{\linewidth}{@{}lXXX@{}}
\toprule
Method & $x$       & $y$       & $\varphi$    \\ \midrule




\multicolumn{4}{l}{\textbf{a) GPS-based Inference:}  $\sigma_x, \sigma_y = 1~\si{\meter}, \sigma_\varphi = 4^\circ$} \\

Attn-based $0\%$ & \SI[detect-weight = true]{0.40}{\metre} & \SI[detect-weight = true]{0.45}{\metre} & \SI[detect-weight = true]{1.6}{\degree} \\

Attn-based $5\%$ & \SI[detect-weight = true]{0.37}{\metre} & \SI[detect-weight = true]{0.41}{\metre} & \SI[detect-weight = true]{1.5}{\degree} \\

Attn-based $50\%$ & \SI[detect-weight = true]{0.31}{\metre} & \SI[detect-weight = true]{0.34}{\metre} & \SI[detect-weight = true]{1.1}{\degree} \\


\multicolumn{4}{l}{\textbf{b) Filter-based Inference:}} \\

Attn-based $0\%$ + EKF & \SI[detect-weight = true]{0.29}{\metre} & \SI[detect-weight = true]{0.33}{\metre} & \SI[detect-weight = true]{1.5}{\degree} \\

Attn-based $5\%$ + EKF& \SI[detect-weight = true]{0.29}{\metre} & \SI[detect-weight = true]{0.31}{\metre} & \SI[detect-weight = true]{1.5}{\degree} \\

Attn-based $50\%$ + EKF& \SI[detect-weight = true]{0.23}{\metre} & \SI[detect-weight = true]{0.25}{\metre} & \SI[detect-weight = true]{0.9}{\degree} \\


 \bottomrule
\end{tabularx}

\end{table}

\subsection{Simulated Experiments}
The results of our simulated experiments are shown in Table~\ref{tab:sim_experiments}. Here, we report the RMSE for the GPS-based inference with noise parameters $\sigma_x, \sigma_y = 1~\si{\meter}, \sigma_\varphi = 4^\circ$ in Table~\ref{tab:sim_experiments}~a), as well as the filter-based inference method in Table~\ref{tab:sim_experiments}~b). For each experiment, we train the network with the synthetic measurements and landmarks, as explained in Section~\ref{sec:sim_framework}, and perform the inference on our Ulm-Lehr dataset. For this experiment, we rely on the Gaussian mixture distribution shown in Fig.~\ref{fig:sim_meas_dist}~c). Furthermore, we also add a small percentage of the real-world samples from our dataset to the training pipeline, indicated as $0\%, 5\%$ and $50\%$. Thus, we demonstrate that a satisfying localization accuracy can be achieved with very few or even no real-world data at all, which still meets our required accuracy in urban scenarios. Especially when the EKF with the CTRV motion model is employed, the negative impact of the synthetic measurement is negligible, thus indicating promising potential for our method to be deployed in a variety of different environments and scenarios around the world.

\section{Conclusion}
We presented a localization approach based on current measurements and generic landmarks from a HD  digital map. Our attention mechanism learns the landmark to measurement association, which we use to register the input point clouds (measurements and landmarks) and, subsequently, infer the vehicle's pose. Furthermore, we proposed a training framework to generate synthetic training data. It facilitates the learning process of our method and improves generalization. Finally, we evaluate our approach on two datasets and show promising results compared to the related work.


\begin{thebibliography}{10}
\providecommand{\url}[1]{#1}
\csname url@rmstyle\endcsname
\providecommand{\newblock}{\relax}
\providecommand{\bibinfo}[2]{#2}
\providecommand\BIBentrySTDinterwordspacing{\spaceskip=0pt\relax}
\providecommand\BIBentryALTinterwordstretchfactor{4}
\providecommand\BIBentryALTinterwordspacing{\spaceskip=\fontdimen2\font plus
\BIBentryALTinterwordstretchfactor\fontdimen3\font minus
  \fontdimen4\font\relax}
\providecommand\BIBforeignlanguage[2]{{%
\expandafter\ifx\csname l@#1\endcsname\relax
\typeout{** WARNING: IEEEtran.bst: No hyphenation pattern has been}%
\typeout{** loaded for the language `#1'. Using the pattern for}%
\typeout{** the default language instead.}%
\else
\language=\csname l@#1\endcsname
\fi
#2}}

\bibitem{thrun2002probabilistic}
S.~Thrun, ``Probabilistic robotics,'' \emph{Communications of the ACM},
  vol.~45, no.~3, pp. 52--57, 2002.

\bibitem{wiederer2020traffic}
J.~Wiederer, A.~Bouazizi, U.~Kressel, and V.~Belagiannis, ``Traffic control
  gesture recognition for autonomous vehicles,'' in \emph{2020 IEEE/RSJ
  International Conference on Intelligent Robots and Systems (IROS)}.\hskip 1em
  plus 0.5em minus 0.4em\relax IEEE, 2020, pp. 10\,676--10\,683.

\bibitem{hasan2019forecasting}
I.~Hasan, F.~Setti, T.~Tsesmelis, V.~Belagiannis, S.~Amin, A.~Del~Bue,
  M.~Cristani, and F.~Galasso, ``Forecasting people trajectories and head poses
  by jointly reasoning on tracklets and vislets,'' \emph{IEEE transactions on
  pattern analysis and machine intelligence}, vol.~43, no.~4, pp. 1267--1278,
  2019.

\bibitem{engel2018deep}
N.~Engel, S.~Hoermann, P.~Henzler, and K.~Dietmayer, ``Deep object tracking on
  dynamic occupancy grid maps using rnns,'' in \emph{2018 21st International
  Conference on Intelligent Transportation Systems (ITSC)}.\hskip 1em plus
  0.5em minus 0.4em\relax IEEE, 2018, pp. 3852--3858.

\bibitem{thrun2001robust}
S.~Thrun, D.~Fox, W.~Burgard, and F.~Dellaert, ``Robust monte carlo
  localization for mobile robots,'' \emph{Artificial intelligence}, vol. 128,
  no. 1-2, pp. 99--141, 2001.

\bibitem{gies2020extended}
F.~{Gies}, J.~{Posselt}, M.~{Buchholz}, and K.~{Dietmayer}, ``Extended
  existence probability using digital maps for object verification,'' in
  \emph{2020 IEEE 23rd International Conference on Information Fusion
  (FUSION)}, 2020, pp. 1--7.

\bibitem{levinson2007map}
J.~Levinson, M.~Montemerlo, and S.~Thrun, ``Map-based precision vehicle
  localization in urban environments.'' in \emph{Robotics: science and
  systems}, vol.~4, no. Citeseer.\hskip 1em plus 0.5em minus 0.4em\relax
  Citeseer, 2007, p.~1.

\bibitem{wing2005consumer}
M.~G. Wing, A.~Eklund, and L.~D. Kellogg, ``Consumer-grade global positioning
  system (gps) accuracy and reliability,'' \emph{Journal of forestry}, vol.
  103, no.~4, pp. 169--173, 2005.

\bibitem{besl1992method}
P.~J. Besl and N.~D. McKay, ``Method for registration of 3-d shapes,'' in
  \emph{Sensor fusion IV: control paradigms and data structures}, vol.
  1611.\hskip 1em plus 0.5em minus 0.4em\relax International Society for Optics
  and Photonics, 1992, pp. 586--606.

\bibitem{montemerlo2002fastslam}
M.~Montemerlo, S.~Thrun, D.~Koller, B.~Wegbreit, \emph{et~al.}, ``Fastslam: A
  factored solution to the simultaneous localization and mapping problem,''
  \emph{Aaai/iaai}, vol. 593598, 2002.

\bibitem{chen2020survey}
C.~Chen, B.~Wang, C.~X. Lu, N.~Trigoni, and A.~Markham, ``A survey on deep
  learning for localization and mapping: Towards the age of spatial machine
  intelligence,'' \emph{arXiv preprint arXiv:2006.12567}, 2020.

\bibitem{engel2019deeplocalization}
N.~Engel, S.~Hoermann, M.~Horn, V.~Belagiannis, and K.~Dietmayer,
  ``Deeplocalization: Landmark-based self-localization with deep neural
  networks,'' in \emph{2019 IEEE Intelligent Transportation Systems Conference
  (ITSC)}.\hskip 1em plus 0.5em minus 0.4em\relax IEEE, 2019, pp. 926--933.

\bibitem{radwan2018vlocnet++}
N.~Radwan, A.~Valada, and W.~Burgard, ``Vlocnet++: Deep multitask learning for
  semantic visual localization and odometry,'' \emph{IEEE Robotics and
  Automation Letters}, vol.~3, no.~4, pp. 4407--4414, 2018.

\bibitem{censi2008icp}
A.~Censi, ``An icp variant using a point-to-line metric,'' in \emph{2008 IEEE
  International Conference on Robotics and Automation}.\hskip 1em plus 0.5em
  minus 0.4em\relax Ieee, 2008, pp. 19--25.

\bibitem{fontanelli2007fast}
D.~Fontanelli, L.~Ricciato, and S.~Soatto, ``A fast ransac-based registration
  algorithm for accurate localization in unknown environments using lidar
  measurements,'' in \emph{2007 IEEE International Conference on Automation
  Science and Engineering}.\hskip 1em plus 0.5em minus 0.4em\relax IEEE, 2007,
  pp. 597--602.

\bibitem{teslic2011ekf}
L.~Tesli{\'c}, I.~{\v{S}}krjanc, and G.~Klan{\v{c}}ar, ``Ekf-based localization
  of a wheeled mobile robot in structured environments,'' \emph{Journal of
  Intelligent \& Robotic Systems}, vol.~62, no.~2, pp. 187--203, 2011.

\bibitem{dellaert1999monte}
F.~Dellaert, D.~Fox, W.~Burgard, and S.~Thrun, ``Monte carlo localization for
  mobile robots,'' in \emph{Proceedings 1999 IEEE International Conference on
  Robotics and Automation (Cat. No. 99CH36288C)}, vol.~2.\hskip 1em plus 0.5em
  minus 0.4em\relax IEEE, 1999, pp. 1322--1328.

\bibitem{montemerlo2003fastslam}
M.~Montemerlo, S.~Thrun, D.~Koller, B.~Wegbreit, \emph{et~al.}, ``Fastslam 2.0:
  An improved particle filtering algorithm for simultaneous localization and
  mapping that provably converges,'' in \emph{IJCAI}, vol.~3, 2003, pp.
  1151--1156.

\bibitem{stubler2015feature}
M.~St{\"u}bler, J.~Wiest, and K.~Dietmayer, ``Feature-based mapping and
  self-localization for road vehicles using a single grayscale camera,'' in
  \emph{2015 IEEE Intelligent Vehicles Symposium (IV)}.\hskip 1em plus 0.5em
  minus 0.4em\relax IEEE, 2015, pp. 267--272.

\bibitem{mullane2011random}
J.~Mullane, B.-N. Vo, M.~D. Adams, and B.-T. Vo, ``A random-finite-set approach
  to bayesian slam,'' \emph{IEEE transactions on robotics}, vol.~27, no.~2, pp.
  268--282, 2011.

\bibitem{deusch2015labeled}
H.~Deusch, S.~Reuter, and K.~Dietmayer, ``The labeled multi-bernoulli slam
  filter,'' \emph{IEEE Signal Processing Letters}, vol.~22, no.~10, pp.
  1561--1565, 2015.

\bibitem{mur2017orb}
R.~Mur-Artal and J.~D. Tard{\'o}s, ``Orb-slam2: An open-source slam system for
  monocular, stereo, and rgb-d cameras,'' \emph{IEEE Transactions on Robotics},
  vol.~33, no.~5, pp. 1255--1262, 2017.

\bibitem{pumarola2017pl}
A.~Pumarola, A.~Vakhitov, A.~Agudo, A.~Sanfeliu, and F.~Moreno-Noguer,
  ``Pl-slam: Real-time monocular visual slam with points and lines,'' in
  \emph{2017 IEEE international conference on robotics and automation
  (ICRA)}.\hskip 1em plus 0.5em minus 0.4em\relax IEEE, 2017, pp. 4503--4508.

\bibitem{mohamed2019survey}
S.~A. Mohamed, M.-H. Haghbayan, T.~Westerlund, J.~Heikkonen, H.~Tenhunen, and
  J.~Plosila, ``A survey on odometry for autonomous navigation systems,''
  \emph{IEEE Access}, vol.~7, pp. 97\,466--97\,486, 2019.

\bibitem{yang2019sanet}
L.~Yang, Z.~Bai, C.~Tang, H.~Li, Y.~Furukawa, and P.~Tan, ``Sanet: Scene
  agnostic network for camera localization,'' in \emph{Proceedings of the
  IEEE/CVF International Conference on Computer Vision}, 2019, pp. 42--51.

\bibitem{kendall2015posenet}
A.~Kendall, M.~Grimes, and R.~Cipolla, ``Posenet: A convolutional network for
  real-time 6-dof camera relocalization,'' in \emph{Proceedings of the IEEE
  international conference on computer vision}, 2015, pp. 2938--2946.

\bibitem{lu2019deepicp}
W.~Lu, G.~Wan, Y.~Zhou, X.~Fu, P.~Yuan, and S.~Song, ``Deepicp: An end-to-end
  deep neural network for 3d point cloud registration,'' \emph{arXiv preprint
  arXiv:1905.04153}, 2019.

\bibitem{wang2020atloc}
B.~Wang, C.~Chen, C.~X. Lu, P.~Zhao, N.~Trigoni, and A.~Markham, ``Atloc:
  Attention guided camera localization,'' in \emph{Proceedings of the AAAI
  Conference on Artificial Intelligence}, vol.~34, no.~06, 2020, pp.
  10\,393--10\,401.

\bibitem{lu2019l3}
W.~Lu, Y.~Zhou, G.~Wan, S.~Hou, and S.~Song, ``L3-net: Towards learning based
  lidar localization for autonomous driving,'' in \emph{Proceedings of the
  IEEE/CVF Conference on Computer Vision and Pattern Recognition}, 2019, pp.
  6389--6398.

\bibitem{qi2017pointnet++}
C.~R. Qi, L.~Yi, H.~Su, and L.~J. Guibas, ``Pointnet++: Deep hierarchical
  feature learning on point sets in a metric space,'' \emph{Advances in neural
  information processing systems}, vol.~30, pp. 5099--5108, 2017.

\bibitem{vaswani2017attention}
A.~Vaswani, N.~Shazeer, N.~Parmar, J.~Uszkoreit, L.~Jones, A.~N. Gomez,
  L.~Kaiser, and I.~Polosukhin, ``Attention is all you need,'' \emph{arXiv
  preprint arXiv:1706.03762}, 2017.

\bibitem{engel2020point}
N.~Engel, V.~Belagiannis, and K.~Dietmayer, ``Point transformer,'' \emph{arXiv
  preprint arXiv:2011.00931}, 2020.

\bibitem{ba2016layer}
J.~L. Ba, J.~R. Kiros, and G.~E. Hinton, ``Layer normalization,'' \emph{arXiv
  preprint arXiv:1607.06450}, 2016.

\bibitem{qi2017pointnet}
C.~R. Qi, H.~Su, K.~Mo, and L.~J. Guibas, ``Pointnet: Deep learning on point
  sets for 3d classification and segmentation,'' in \emph{Proceedings of the
  IEEE conference on computer vision and pattern recognition}, 2017, pp.
  652--660.

\bibitem{belagiannis2015robust}
V.~Belagiannis, C.~Rupprecht, G.~Carneiro, and N.~Navab, ``Robust optimization
  for deep regression,'' in \emph{Proceedings of the IEEE international
  conference on computer vision}, 2015, pp. 2830--2838.

\bibitem{kendall2018multi}
A.~Kendall, Y.~Gal, and R.~Cipolla, ``Multi-task learning using uncertainty to
  weigh losses for scene geometry and semantics,'' in \emph{Proceedings of the
  IEEE conference on computer vision and pattern recognition}, 2018, pp.
  7482--7491.

\bibitem{ester1996density}
M.~Ester, H.-P. Kriegel, J.~Sander, X.~Xu, \emph{et~al.}, ``A density-based
  algorithm for discovering clusters in large spatial databases with noise.''
  in \emph{Kdd}, vol.~96, no.~34, 1996, pp. 226--231.

\bibitem{matas2004robust}
J.~Matas, O.~Chum, M.~Urban, and T.~Pajdla, ``Robust wide-baseline stereo from
  maximally stable extremal regions,'' \emph{Image and vision computing},
  vol.~22, no.~10, pp. 761--767, 2004.

\bibitem{stubler2017continuously}
M.~St{\"u}bler, S.~Reuter, and K.~Dietmayer, ``A continuously learning
  feature-based map using a bernoulli filtering approach,'' in \emph{2017
  Sensor Data Fusion: Trends, Solutions, Applications (SDF)}.\hskip 1em plus
  0.5em minus 0.4em\relax IEEE, 2017, pp. 1--6.

\bibitem{Geiger2012CVPR}
A.~Geiger, P.~Lenz, and R.~Urtasun, ``Are we ready for autonomous driving? the
  kitti vision benchmark suite,'' in \emph{Conference on Computer Vision and
  Pattern Recognition (CVPR)}, 2012.

\bibitem{paszke2019pytorch}
A.~Paszke, S.~Gross, F.~Massa, A.~Lerer, J.~Bradbury, G.~Chanan, T.~Killeen,
  Z.~Lin, N.~Gimelshein, L.~Antiga, \emph{et~al.}, ``Pytorch: An imperative
  style, high-performance deep learning library,'' \emph{arXiv preprint
  arXiv:1912.01703}, 2019.

\bibitem{mullane2010rao}
J.~Mullane, B.-N. Vo, and M.~D. Adams, ``Rao-blackwellised phd slam,'' in
  \emph{2010 IEEE International Conference on Robotics and Automation}.\hskip
  1em plus 0.5em minus 0.4em\relax IEEE, 2010, pp. 5410--5416.

\end{thebibliography}

\end{document}